\documentclass{article}

\usepackage{amsmath}
\usepackage{graphicx}
\usepackage{caption}
\usepackage{amsfonts}
\usepackage{booktabs}
\usepackage{pifont}

\usepackage{wrapfig}
\usepackage{multirow}
\usepackage{hyperref}
\hypersetup{
    colorlinks=true,
    linkcolor=blue!55!black,
    citecolor=blue!55!black,
    urlcolor=blue!55!black,
}
\usepackage[preprint]{corl_2026}

\title{You Only Touch Once: 6-DoF Object Pose Estimation from Single Tactile Contact}

\author{
  \textbf{Pengfei Ye}$^{1\dagger}$\enspace
  \textbf{Yuxiang Ma}$^{1\dagger}$\enspace
  \textbf{Haonan Chen}$^{2}$\\[3pt]
  \textbf{Guangming Wang}$^{3}$\enspace
  \textbf{Yixiong Jing}$^{3}$\enspace
  \textbf{Brian Sheil}$^{3}$\enspace
  \textbf{Yilun Du}$^{2}$\enspace
  \textbf{Edward Adelson}$^{1\ast}$\\[7pt]
  \small
  $^{1}$MIT CSAIL\quad
  $^{2}$Harvard University\quad
  $^{3}$University of Cambridge\\[2pt]
  \normalsize
  \textsuperscript{$\dagger$}These authors contributed equally to this work.\quad
  \textsuperscript{$\ast$}Corresponding author.
}

\begin{document}
\maketitle

\begin{figure}[h]
    \vspace{-25pt}
    \centering
    \includegraphics[width=\linewidth]{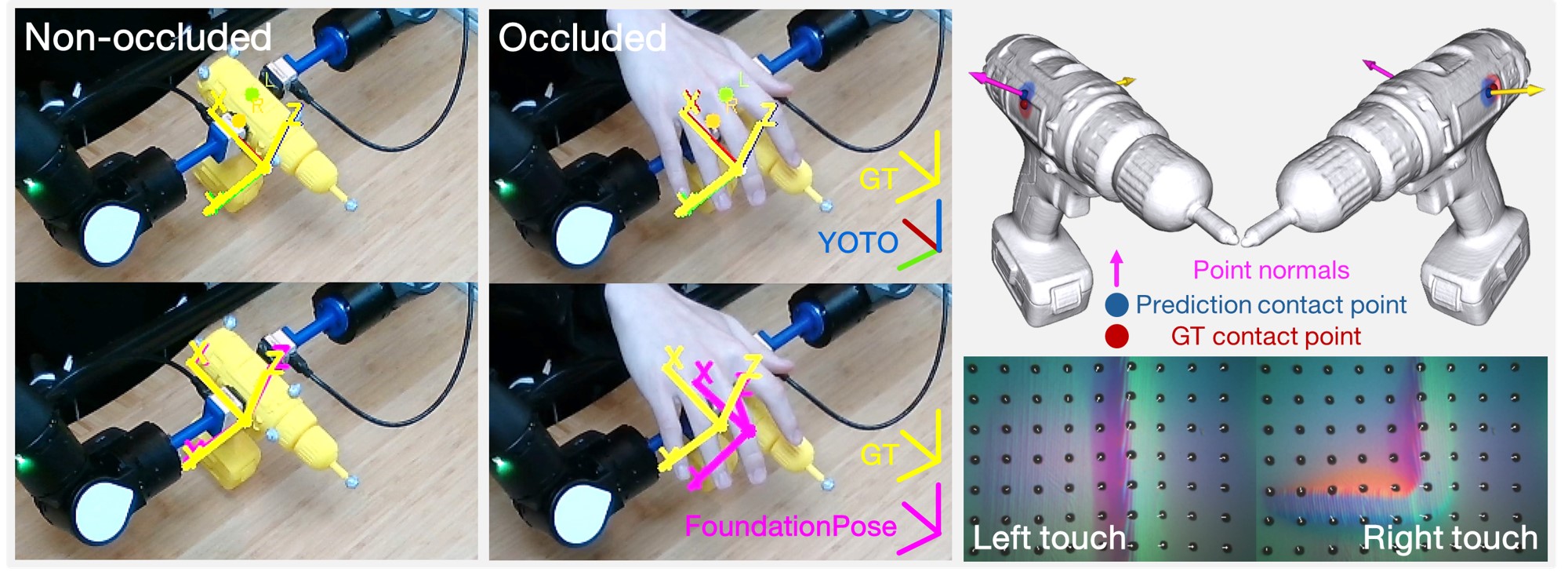}
    \captionsetup{font=footnotesize}
    \caption{YOTO recovers 6-DoF object pose from a single dual-GelSight 
    contact. \textbf{Left/middle}: drill under non-occluded and occluded 
    views; YOTO (top, RGB axes) remains anchored to the object while 
    FoundationPose (bottom, magenta) drifts onto the occluder or a wrong 
    canonical viewpoint. Mocap GT axes in yellow. \textbf{Right}: method 
    illustration on a scanned drill, showing predicted contacts (blue) 
    vs.\ GT contact regions (red), kNN-aggregated surface normals 
    (magenta arrows), and raw GelSight images.}
    \label{fig:teaser}
    \vspace{-10pt}
\end{figure}


\begin{abstract}
    Accurate 6-DoF object pose estimation is fundamental to robotic manipulation, yet vision-based methods often fail under occlusion, poor lighting, and reflective or transparent surfaces. We present YOTO, a tactile-only pose estimation system that recovers the full 6-DoF object pose from a single pair of simultaneous contacts, without requiring contact history. YOTO represents each tactile contact as a local 3D point cloud and localizes it on the object surface through a coarse-to-fine network. The two localized contacts, together with the calibrated sensor poses, are then fed to a closed-form normal-aware SVD solver that recovers the full 6-DoF object pose in one step. To reduce real-data requirements, the localization network is pretrained on virtual tactile patches sampled from the object model and fine-tuned with a small number of real contacts. We further show that YOTO can operate on object models reconstructed from consumer-grade mobile scans, and quantify the gap relative to CAD-based models. Experiments on four geometrically diverse objects demonstrate accurate tactile contact localization and pose estimation, outperforming vision-based and geometric baselines, especially when visual perception is unreliable. Code, trained models, and the real GelSight dataset will be released upon publication.
\end{abstract}



\section{Introduction}

    Reliable 6-DoF object pose estimation underpins virtually every 
    contact-rich manipulation task, from grasping and placement to in-hand 
    reorientation and tool use. The dominant paradigm pairs RGB or RGB-D 
    sensing with learned pose estimators~\cite{Kendall_2015_ICCV,peng2019pvnet,
    wen2024foundationpose,ornek2024foundpose}, which achieve impressive 
    accuracy when the object is clearly visible to the camera.
    
    However, these methods degrade precisely in the regimes where 
    manipulation most needs them: the manipulator's own hand or gripper 
    occludes the object during grasping; transparent, specular, or 
    low-textured surfaces violate photometric assumptions; and uneven or 
    dim lighting corrupts the input. Tactile sensing observes geometry only at the point of contact and is robust to all three failure modes, but introduces a complementary challenge.
    
    Tactile pose estimation faces a severe localization ambiguity. A single tactile contact observes only a small surface patch, while the same local geometry may appear at many locations on the object. Existing tactile approaches either match low-dimensional contact representations~\cite{bauza2023tac2pose} or rely on contact history for tracking~\cite{10766628}, limiting single-shot pose estimation on geometrically complex objects.
    
    We present YOTO, a tactile-only system that recovers absolute 6-DoF object pose from one simultaneous dual-GelSight contact. YOTO represents each contact as a 3D point cloud and localizes it on the object surface using a coarse-to-fine network. The object surface is partitioned into block point clouds, reducing global contact localization to retrieval over local surface regions, followed by offset regression within candidate blocks. The two localized contacts, together with kNN-aggregated surface normals estimated from the parent cloud around each predicted contact, and calibrated sensor poses, are then used in a closed-form normal-aware SVD solver.
    
    A central practical concern for tactile pose estimation is the cost of 
    collecting real GelSight data, and the typical reliance on hand-crafted CAD meshes of each target object. We address both with a two-stage training pipeline: virtual contact patches automatically sampled from the object model provide dense pretraining supervision, while a small set of real GelSight contacts is used for fine-tuning. Crucially, the pipeline operates on object models reconstructed from consumer-grade 3D scanners, removing the CAD-mesh prerequisite that prior tactile methods typically assume; we quantify the small accuracy gap relative to CAD-trained counterparts in Sec.~\ref{sec:exp_localization}.
    
    \noindent Our contributions are:
    
    \textbf{System.} YOTO recovers absolute 6-DoF pose from a single dual-sensor tactile contact on geometrically complex objects, without requiring contact history, object motion, or visual input.
    
    \textbf{Method.} A surface block representation enables coarse-to-fine 3D tactile contact localization, which is coupled with a normal-aware dual-contact SVD pose solver.
    
    \textbf{Data pipeline.} An automatic virtual contact generation pipeline with two-stage training achieves sim-to-real transfer with limited physical data.
    
    \textbf{Benchmark.} Evaluation on four objects, using both CAD and consumer-grade scan models, demonstrates competitive performance against vision-based and geometric baselines, especially in visually degraded settings.
    
\vspace{-10pt}
\section{Related Work}
    \paragraph{Vision-based and vision-tactile fusion pose estimation.}
    Vision-based 6-DoF pose estimation spans a spectrum from template-based 
    matching~\cite{hinterstoisser2012model}, direct pose 
    regression~\cite{Kendall_2015_ICCV}, implicit pose 
    embeddings~\cite{sundermeyer2018implicit}, and keypoint voting~\cite{peng2019pvnet} 
    to recent foundation-model approaches that generalize to novel objects 
    from a single CAD reference~\cite{wen2024foundationpose,ornek2024foundpose}, 
    achieving accuracy within a few millimeters and degrees on textured, 
    well-lit objects. Learned dense local descriptors~\cite{Zeng_2017_CVPR,Choy_2019_ICCV} 
    have likewise advanced 3D feature matching in point-cloud space, a 
    paradigm we adopt for tactile patches in Sec.~\ref{sec:localization_network}. These methods can achieve strong performance when RGB or RGB-D observations are reliable, but manipulation often violates this assumption through hand occlusion, poor lighting, specular reflection, or transparent objects. Vision-tactile fusion methods address this by combining external cameras with tactile sensing for in-hand pose estimation, shape reconstruction, or neural-field-based perception~\cite{dikhale2022visuotactile,suresh2022shapemap,suresh2024neuralfeels}. In contrast, YOTO targets tactile-only absolute pose estimation from a single dual-contact observation.
    
    \paragraph{Tactile pose estimation from contact observations.}
    Estimating object pose from local tactile contacts is challenging because small contact patches can be geometrically ambiguous. Early work accumulated contact evidence probabilistically over time~\cite{petrovskaya2011global}, while Tac2Pose learns object-specific embeddings over simulated contact shapes and matches a tactile observation against a library of pose hypotheses~\cite{bauza2023tac2pose,pmlr-v155-villalonga21a}. Closely related to multi-contact sensing, Caddeo et al.~\cite{caddeo2023collision} sample candidate contacts on a known mesh and refine poses using tactile image features, geometric constraints, and optimization. YOTO instead represents each tactile contact as a 3D point cloud, localizes it to a surface block of the parent model, and recovers pose analytically from two localized contacts, associated surface normals, and calibrated sensor poses.
    
    \paragraph{Tactile tracking and temporal contact reasoning.}
    Several tactile systems reduce ambiguity by using motion or temporal contact histories. PatchGraph~\cite{sodhi2022patchgraph} builds a tactile patch graph for in-hand tracking, MidasTouch~\cite{suresh2023midastouch} performs particle filtering over sliding touch, and NormalFlow~\cite{10766628} tracks object pose from tactile flow. These methods exploit sequential contacts, sliding motion, or an initial tracking state. YOTO addresses a complementary setting by estimating the absolute 6-DoF pose from one simultaneous dual-sensor contact, without sliding, contact history, or an initial pose estimate.


\section{From Tactile Contacts to 6-DoF Pose}
\label{sec:method}

    \begin{figure}[h]
        \vspace{-10pt}
        \centering
        \includegraphics[width=0.95\linewidth]{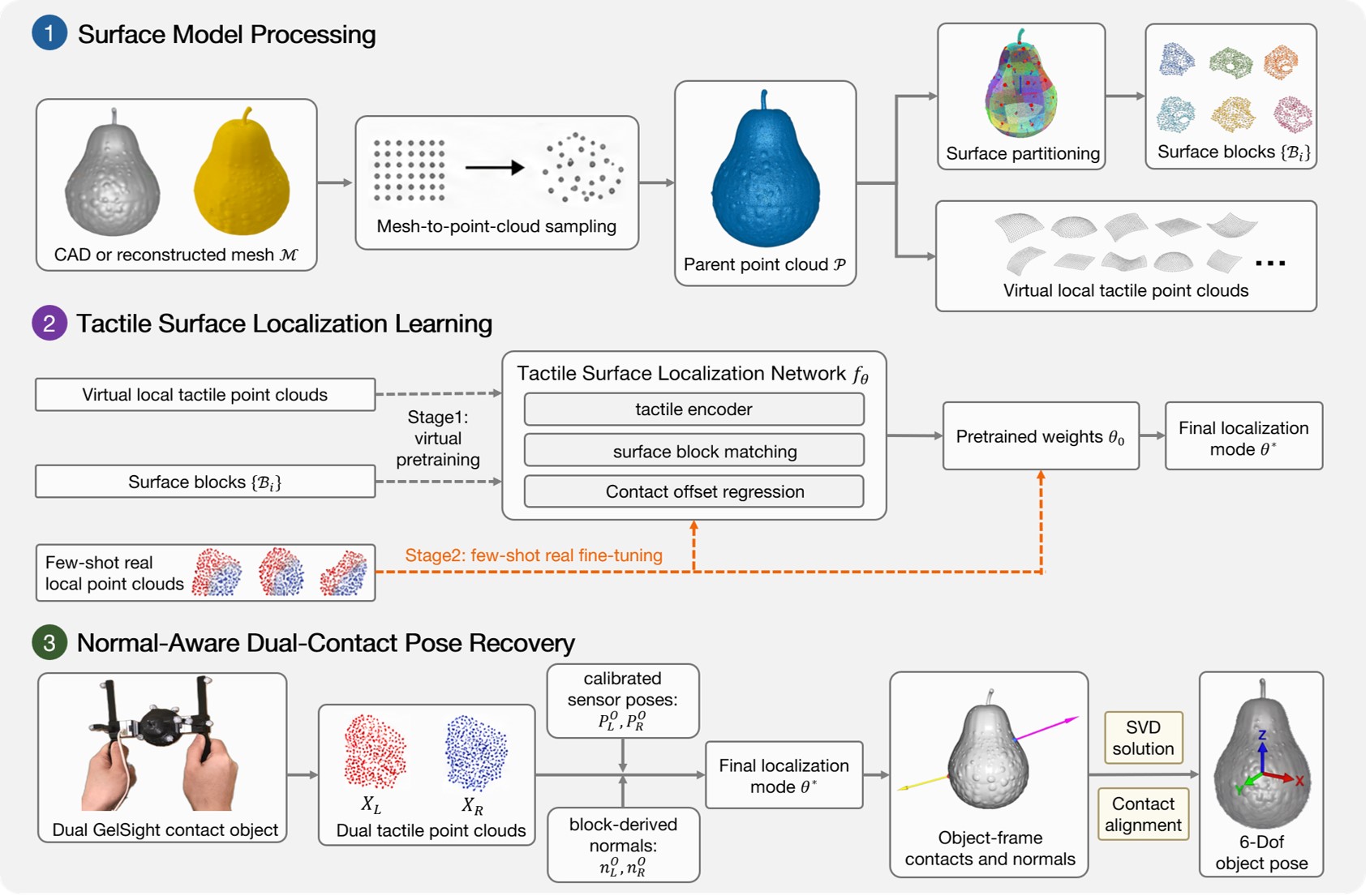}
        \captionsetup{font=footnotesize}
        \caption{YOTO system pipeline (Sec.~\ref{sec:method}): surface representation, coarse-to-fine localization with virtual pretraining and few-shot real fine-tuning, and a closed-form normal-aware SVD pose solver.}
        \label{fig:pipeline}
        \vspace{-10pt}
    \end{figure}

    Given an object model $\mathcal{M}$ and a simultaneous dual-GelSight contact, YOTO outputs the 6-DoF pose $\hat{\mathbf{T}}_O^W\in SE(3)$. Fig.~\ref{fig:pipeline} shows the pipeline: surface representation (Sec.~\ref{sec:surface_representation}), coarse-to-fine localization (Sec.~\ref{sec:localization_network}--\ref{sec:training_objective}), and SVD pose recovery (Sec.~\ref{sec:pose_recovery}).

\subsection{Surface Representation and Virtual Contact Generation}
\label{sec:surface_representation}

    YOTO samples a parent surface point cloud $\mathcal{P}=\{(\mathbf{x}_j,\mathbf{n}_j)\}_{j=1}^{N}$ from $\mathcal{M}$, where $\mathbf{x}_j$ and $\mathbf{n}_j$ are object-frame surface points and normals. The input can be either a high-fidelity CAD mesh or a reconstructed mesh from commodity 3D scanning tools; both are processed by the same point-cloud pipeline, enabling training from standard 3D scanning or surface reconstruction outputs~\cite{6162880,Schonberger_2016_CVPR,kazhdan2006poisson}.
    
    To reduce the search space, we partition $\mathcal{P}$ into a grid of surface blocks $\{\mathcal{B}_i\}_{i=1}^{N_B}$ aligned with the object's principal axes. Each block stores its surface points and normals, an object-frame center $\mathbf{c}_i^O$. This turns dense matching over the full surface into block retrieval followed by local offset regression.
    
    Virtual tactile point clouds are sampled from Euclidean-ball neighbourhoods of $\mathcal{P}$ around randomly chosen centres, with the ball radius matched to a GelSight footprint. Each virtual sample is assigned a dominant block index $i^\star$, an object-frame contact center $\mathbf{p}^O$, and an offset $\Delta\mathbf{p}^{O}=\mathbf{p}^{O}-\mathbf{c}_{i^\star}^{O}$, which supervise coarse retrieval and fine regression.

    To make virtual and real patches feature-compatible, every patch
    fed to the network, whether sampled from $\mathcal{P}$ during
    training or reconstructed from a real GelSight contact at test
    time, is first lifted into a common sensor-aligned frame. For
    each patch we estimate the local concave direction via PCA on
    the patch coordinates, rotate the patch so the concave direction
    aligns with the positive $z$ axis, and translate it so the
    geometric centroid sits at the origin. The canonicalized
    coordinates are fed to the network, while the patch's original
    object-frame location $\mathbf{p}^O$ remains recorded as the
    regression target. Without this step, virtual patches would
    carry their absolute orientation on the object as part of their
    input. The network would then memorize per-location orientation
    cues as a shortcut to the ground-truth target rather than learn
    the local contact geometry. Real GelSight patches, on the other
    hand, arrive natively in the sensor's local frame, with the gel
    surface on $xy$ and the contact normal along $z$. They would
    therefore occupy a visibly different feature distribution from
    object-frame virtual patches at test time. Canonicalization
    removes both effects and enables the few-shot sim-to-real
    transfer in Sec.~\ref{sec:training_objective}.

\vspace{-5pt}
\subsection{Coarse-to-Fine Contact Localization Network}
\label{sec:localization_network}

    \begin{figure*}[t]
        \centering
        \includegraphics[width=\textwidth]{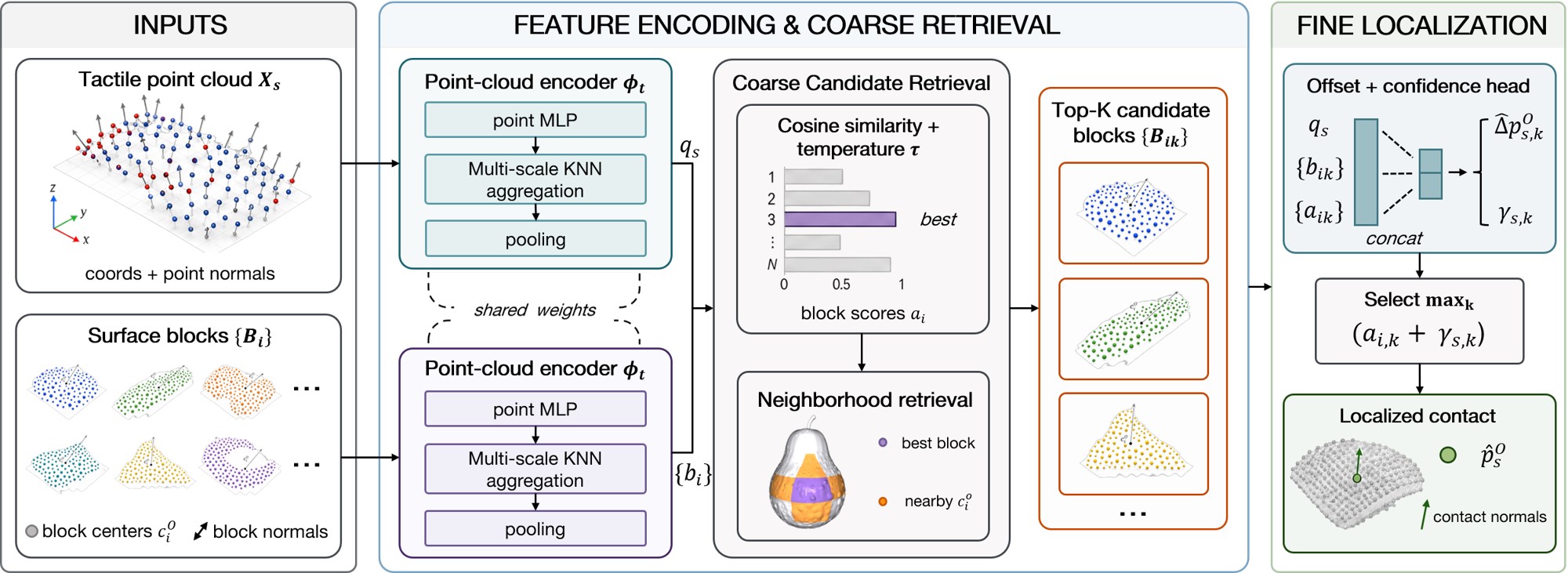}
        \captionsetup{font=footnotesize}
        \caption{Coarse-to-fine tactile surface localization network (Sec.~\ref{sec:localization_network}).}
        \label{fig:localization_network}
        \vspace{-15pt}
    \end{figure*}
    
    Given a tactile point cloud $\mathbf{X}_s$ from sensor $s \in \{L, R\}$ (left and right GelSight) and the surface blocks $\{\mathcal{B}_i\}$ defined in Sec.~\ref{sec:surface_representation}, the localization network predicts the 3D contact position $\hat{\mathbf{p}}_s^O$ in the object frame through the coarse-to-fine pipeline shown in Fig.~\ref{fig:localization_network}. The tactile patch and surface blocks are processed by two weight-sharing branches of a lightweight point-cloud encoder~\cite{Qi_2017_CVPR,NIPS2017_d8bf84be} that take coordinates and per-point normals as inputs, mapping both into a common feature space. We denote the tactile feature as $\mathbf{q}_s=\phi_t(\mathbf{X}_s)$ and the block feature as $\mathbf{b}_i=\phi_b(\mathcal{B}_i)$.
    
    The coarse stage, illustrated in the middle panel of Fig.~\ref{fig:localization_network}, scores each surface block by normalized feature similarity,
    \begin{equation}
        a_i
        =
        \tau
        \frac{
        \mathbf{q}_s^\top \mathbf{b}_i
        }{
        \|\mathbf{q}_s\|_2 \|\mathbf{b}_i\|_2
        },
    \end{equation}
    where $\tau$ is a learnable temperature. The highest-scoring block gives an initial contact region. YOTO then retrieves a spatial neighborhood around this block using the object-frame centers $\mathbf{c}_i^O$ and keeps the top-$K$ candidates, retaining multiple nearby hypotheses for fine localization.
    
    For each candidate block $\mathcal{B}_{i_k}$, the fine localization head shown on the right of Fig.~\ref{fig:localization_network} predicts a local offset and a confidence score,
    \begin{equation}
        (\Delta\hat{\mathbf{p}}_{s,k}^O, \gamma_{s,k})
        =
        g_\theta(\mathbf{q}_s,\mathbf{b}_{i_k}),
        \qquad
        \hat{\mathbf{p}}_{s,k}^O
        =
        \mathbf{c}_{i_k}^O
        +
        \Delta\hat{\mathbf{p}}_{s,k}^O .
    \end{equation}
    The final prediction is selected by combining the coarse matching score and the predicted confidence,
    \begin{equation}
        k^\star
        =
        \arg\max_k
        \left(
        a_{i_k} + \gamma_{s,k}
        \right),
        \qquad
        \hat{\mathbf{p}}_s^O
        =
        \hat{\mathbf{p}}_{s,k^\star}^O .
    \end{equation}
    After localization, YOTO estimates the contact-point surface normal $\hat{\mathbf{n}}_s^O$ by averaging the parent-cloud normals of the $k$ nearest surface points to $\hat{\mathbf{p}}_s^O$ ($k{=}20$ in our experiments). This kNN-aggregated normal is more locally faithful than a single block-wide normal, and is passed with the predicted contact position to the pose solver in Sec.~\ref{sec:pose_recovery}.

\vspace{-5pt}
\subsection{Two-Stage Training Objective}
\label{sec:training_objective}

    YOTO is trained in two stages, mirroring sim-to-real practice in 
    vision~\cite{8202133} and tactile sensing~\cite{9681378}. First, the 
    localization network is pretrained on virtual tactile point clouds 
    sampled from known surface locations, providing supervision for both 
    contact block retrieval and object-frame contact position regression. Second, the network is fine-tuned with a mixture of virtual samples and real GelSight contacts. Virtual samples preserve dense surface coverage, while real samples adapt the model to sensing artifacts and residual simulation-to-real differences. Real samples receive a larger loss weight during fine-tuning.
    
    For each training sample, the ground-truth contact position $\mathbf{p}_s^O$ defines a ground-truth block label $y_s$ and an offset from the corresponding block center,
    \begin{equation}
        \Delta \mathbf{p}_s^O
        =
        \mathbf{p}_s^O - \mathbf{c}_{y_s}^O ,
    \end{equation}
    where $\mathbf{c}_{y_s}^O$ is the object-frame center of the ground-truth block $\mathcal{B}_{y_s}$. The coarse retrieval stage in Fig.~\ref{fig:localization_network} is supervised with a block classification loss,
    \begin{equation}
        \mathcal{L}_{\mathrm{cls}}
        =
        \mathrm{CE}(\mathbf{a}, y_s),
    \end{equation}
    where $\mathbf{a}$ denotes the block matching logits. To keep the correct block competitive during candidate selection, we also use a margin loss over the highest-scoring block and the retained top-$K$ candidates,
    \begin{equation}
        \mathcal{L}_{\mathrm{topK}}
        =
        \left[
        a_{\hat{i}} - a_{y_s} + m_1
        \right]_+
        +
        \left[
        a_{i_K} - a_{y_s} + m_2
        \right]_+ ,
    \end{equation}
    where $\hat{i}$ is the highest-scoring block, $i_K$ is the lowest-scoring block among the retained top-$K$ candidates, and $[\cdot]_+=\max(\cdot,0)$.
    
    The fine stage is supervised by a smooth-$\ell_1$ regression loss over candidate contact positions,
    \begin{equation}
        \mathcal{L}_{\mathrm{pos}}
        =
        \sum_{k=1}^{K}
        w_k\,
        \mathrm{SmoothL1}
        \left(
        \hat{\mathbf{p}}_{s,k}^O,
        \mathbf{p}_s^O
        \right),
    \end{equation}
    where $w_k$ favors candidates with higher combined matching confidence and closer block centers. The full objective is
    \begin{equation}
        \mathcal{L}
        =
        \lambda_{\mathrm{c}}
        \mathcal{L}_{\mathrm{cls}}
        +
        \lambda_{\mathrm{k}}
        \mathcal{L}_{\mathrm{topK}}
        +
        \lambda_{\mathrm{p}}
        \mathcal{L}_{\mathrm{pos}} .
    \end{equation}
    At inference time, the trained localization network predicts $\hat{\mathbf{p}}_s^O$ for each tactile sensor independently. These contact positions and the kNN-aggregated surface normals computed around each $\hat{\mathbf{p}}_s^O$ (Sec.~\ref{sec:localization_network}) are passed to the pose solver below.

\vspace{-5pt}
\subsection{6-DoF Pose Recovery from Dual Contacts}
\label{sec:pose_recovery}

    The localization network of
    Sec.~\ref{sec:localization_network} predicts only a contact
    position $\hat{\mathbf{p}}_s^O$, not a full 6-DoF pose.
    This is a direct consequence of the patch canonicalization in
    Sec.~\ref{sec:surface_representation}. By rotating every patch
    into a common sensor-aligned frame, the network deliberately
    discards information about the patch's absolute orientation on
    the object, since that orientation cannot be inferred
    from a single contact alone. Recovering the missing rotation
    therefore requires an additional geometric constraint, which
    YOTO obtains by pairing the two simultaneous contacts. The
    relative geometry between the two localized contact points and
    their associated surface normals is sufficient to pin down the
    object's full 6-DoF pose in closed form, which we derive below.
    
    The trained localization network produces, for each sensor $s \in \{L, R\}$, an object-frame contact position $\hat{\mathbf{p}}_s^O$ and a locally kNN-aggregated outward surface normal $\hat{\mathbf{n}}_s^O$ computed from the parent cloud around $\hat{\mathbf{p}}_s^O$. YOTO converts these tactile observations into an SE(3) object pose through a closed-form solver. The retrieved normals are essential because two point correspondences alone constrain the contact-to-contact axis but leave the rotation around that axis ambiguous.
    
    Let $\mathbf{p}_L^W$ and $\mathbf{p}_R^W$ denote the corresponding contact centers in the world frame, computed from the calibrated left and right GelSight poses. We form the object-frame and world-frame displacement directions between the two contacts,
    \begin{equation}
        \mathbf{d}^O
        =
        \frac{\hat{\mathbf{p}}_R^O-\hat{\mathbf{p}}_L^O}
        {\|\hat{\mathbf{p}}_R^O-\hat{\mathbf{p}}_L^O\|_2},
        \qquad
        \mathbf{d}^W
        =
        \frac{\mathbf{p}_R^W-\mathbf{p}_L^W}
        {\|\mathbf{p}_R^W-\mathbf{p}_L^W\|_2}.
    \end{equation}
    Together with the object-frame surface normals and the world-frame sensor contact directions, these define two sets of corresponding directions,
    \begin{equation}
        \mathbf{Q}^O
        =
        \begin{bmatrix}
        \mathbf{d}^O &
        \hat{\mathbf{n}}_L^O &
        \hat{\mathbf{n}}_R^O
        \end{bmatrix},
        \qquad
        \mathbf{Q}^W
        =
        \begin{bmatrix}
        \mathbf{d}^W &
        \mathbf{n}_L^W &
        \mathbf{n}_R^W
        \end{bmatrix}.
    \end{equation}
    Here $\mathbf{n}_L^W$ and $\mathbf{n}_R^W$ are obtained from the calibrated sensor orientations and indicate the world-frame outward contact directions. This is analogous to oriented point-pair constraints~\cite{drost2010model} 
    and complements learned point-cloud registration~\cite{Wang_2019_ICCV}, 
    here driven by tactile rather than depth correspondences obtained from 
    sparse contact.
    
    The object rotation is estimated by solving the orthogonal Procrustes problem
    \begin{equation}
        \hat{\mathbf{R}}
        =
        \arg\min_{\mathbf{R}\in SO(3)}
        \left\|
        \mathbf{Q}^W - \mathbf{R}\mathbf{Q}^O
        \right\|_F^2,
    \end{equation}
    which admits a closed-form SVD solution~\cite{arun1987least,umeyama2002least}. Given $\hat{\mathbf{R}}$, the translation is computed by aligning the two predicted object-frame contact points to their world-frame counterparts,
    \begin{equation}
        \hat{\mathbf{t}}
        =
        \frac{1}{2}
        \sum_{s\in\{L,R\}}
        \left(
        \mathbf{p}_s^W - \hat{\mathbf{R}}\hat{\mathbf{p}}_s^O
        \right).
    \end{equation}
    The final estimated object pose is $\hat{\mathbf{T}}_O^W=[\hat{\mathbf{R}},\hat{\mathbf{t}}]$.


\section{Experiments}
\label{sec:experiments}

    We evaluate YOTO on four tabletop objects of contrasting geometry (a drill, a squirrel, a monkey, and an avocado), using two 6-DoF AIRBOT Play arms, each carrying a GelSight Mini sensor~\cite{s17122762} on a rigid probe. The dual-arm configuration lets the two sensors contact the object from independently controlled directions, avoiding the near-coaxial geometry of a parallel gripper that makes dual-contact pose recovery ill-conditioned. An OptiTrack motion-capture system provides ground-truth poses for the object and both sensors and is used only for evaluation. Each GelSight contact is converted to a local point cloud with the InvariantCloud reconstruction pipeline~\cite{ye2026invariantcloud}; the corresponding object-frame ground-truth contact location is computed automatically from the synchronized mocap trajectory and the object mesh. For each object, YOTO is pretrained on virtual contacts and fine-tuned on a small number of real GelSight samples. No prior tactile pose estimator is publicly reproducible under the same single-shot, dual-contact, no-history setting, so we compare against geometry-only ICP as a tactile lower bound and FoundationPose~\cite{wen2024foundationpose} as a representative visual baseline.
    
    
    \subsection{Tactile Contact Localization Accuracy}
    \label{sec:exp_localization}
    
\begin{figure}[!h]
        \centering
        \includegraphics[width=0.8\linewidth]{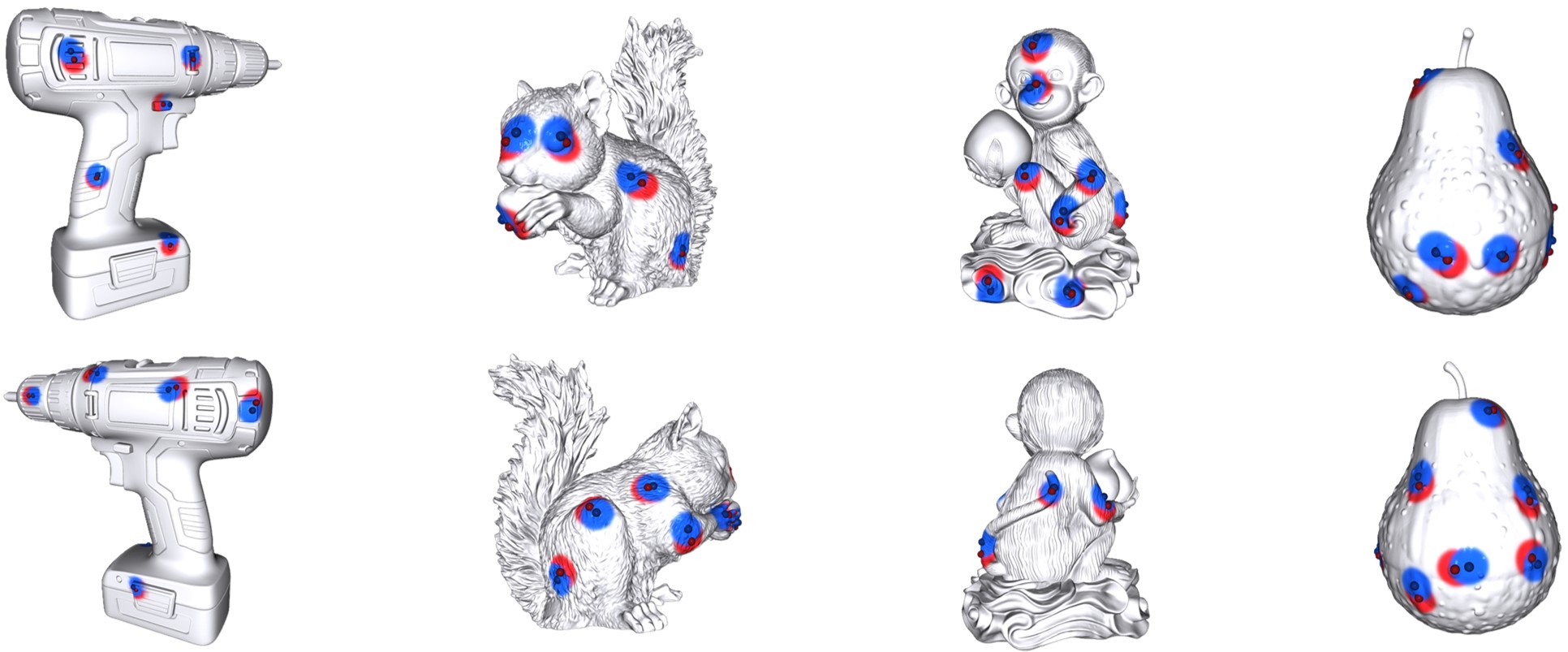}
        \captionsetup{font=footnotesize}
        \caption{Representative tactile contact predictions (blue) vs.\
        ground-truth contact regions (red) for the four test objects, two
        views per object.}
        \label{fig:exp1_inset}
    \end{figure}

    We first evaluate the localization network of Sec.~\ref{sec:localization_network} in isolation. For each object we collect $10$ real GelSight contacts at locations distributed across the surface, then compute the Euclidean distance in the object frame between the predicted contact $\hat{\mathbf{p}}_s^O$ and the mocap-derived ground truth (used only for evaluation, not as model input). Fig.~\ref{fig:exp1_inset} shows representative predictions and ground-truth contact regions overlaid on the four test objects. Tab.~\ref{tab:exp1} reports per-object errors and compares YOTO against a geometry-only ICP baseline and two ablations.
    
    \begin{table}[t]
        \centering
        \captionsetup{font=footnotesize}
        \caption{Per-object tactile contact localization error (mm, lower is better). YOTO uses virtual pretraining followed by real fine-tuning on scanned object models unless noted; the \emph{Scanned, virtual only} row skips the real-data fine-tuning stage.}
        \label{tab:exp1}
        \small
        \setlength{\tabcolsep}{4pt}
        \resizebox{\textwidth}{!}{
        \begin{tabular}{l c c c c c}
        \toprule
        Method                       & Drill                       & Squirrel                    & Monkey                      & Avocado                     & All  \\
        \midrule
        ICP (geometry only)          & $64.36\!\pm\!35.10$         & $47.41\!\pm\!17.16$         & $51.31\!\pm\!15.75$         & $48.17\!\pm\!17.54$         & $52.81\!\pm\!22.98$    \\
        YOTO (Scanned, virtual only) & $28.52\!\pm\!10.75$         & $22.45\!\pm\!7.47$          & $20.90\!\pm\!9.67$          & $26.74\!\pm\!11.63$         & $24.65\!\pm\!10.11$    \\
        YOTO (CAD mesh)               & $5.10\!\pm\!2.07$           & $5.43\!\pm\!1.97$           & $4.05\!\pm\!2.74$           & $7.22\!\pm\!1.56$           & $5.45\!\pm\!2.35$ \\
        \textbf{YOTO (Scanned mesh)} & $\mathbf{3.96\!\pm\!1.38}$  & $\mathbf{4.81\!\pm\!2.16}$  & $\mathbf{4.36\!\pm\!2.38}$  & $\mathbf{5.97\!\pm\!2.54}$  & $\mathbf{4.78\!\pm\!2.21}$ \\
        \bottomrule
        \end{tabular}}
        \vspace{-10pt}
    \end{table}
    
    YOTO achieves a sub-centimeter mean localization error across 40 contacts (Tab.~\ref{tab:exp1}), with a worst case below 10\,mm. The drill, whose elongated body and prominent trigger provide distinctive local geometry, yields the lowest mean error. The avocado provides the greatest challenge because its surface bumps are locally similar to each other; this occasionally drives the coarse retrieval stage into a wrong block from which fine regression cannot recover. Removing the real-data fine-tuning stage raises mean error more than five-fold (Tab.~\ref{tab:exp1}), confirming that a small set of real GelSight samples is necessary to bridge the residual sim-to-real gap. Training on consumer-grade scanned models matches or exceeds CAD-mesh training, as scanned meshes implicitly capture 3D-printing artifacts (layer thickness, surface roughness) that the GelSight actually observes. We use scanned models throughout. In addition to better accuracy,
    YOTO runs roughly $30\times$ faster than the multi-start ICP
    baseline per contact and with substantially smaller variance
    across objects; per-object timing is reported in
    Tab.~\ref{tab:runtime} (Appendix~\ref{app:impl_inference}).

    \subsection{6-DoF Object Pose Estimation under Occlusion}
    \label{sec:exp_pose}

    We next compare the full pose recovery against FoundationPose~\cite{wen2024foundationpose} on the four objects under two viewing conditions: a clear view in which the overhead camera observes the full object, and an occluded view in which the operator's hand covers more than half of the object's projected area. Ground-truth poses come from OptiTrack and are used only for evaluation. For each object/condition pair we evaluate both methods on five distinct object placements; each placement is a continuous recording during which the estimator produces a stream of pose predictions, and we report the time-averaged error per placement. Fig.~\ref{fig:teaser} shows representative captures for both methods.
    
    \begin{table}[ht]
    \centering
    \captionsetup{font=footnotesize}
    \caption{6-DoF pose estimation error (mean $\pm$ std over 5 trials per object, lower is better). YOTO uses tactile contact only; FoundationPose uses the calibrated RGB-D stream.}
    \label{tab:exp2}
    \small
    \setlength{\tabcolsep}{4pt}
    \resizebox{\textwidth}{!}{
    \begin{tabular}{l l c c c c c}
    \toprule
    \multicolumn{7}{c}{\textbf{Translation error (mm)}} \\
    \midrule
    Method & Condition & Drill & Squirrel & Monkey & Avocado & All \\
    \midrule
    \multirow{2}{*}{FoundationPose}
     & clear    & $5.86\!\pm\!0.97$        & $5.93\!\pm\!0.26$        & $6.24\!\pm\!0.53$        & $3.88\!\pm\!0.49$        & $5.48\!\pm\!1.12$ \\
     & occluded & $102.36\!\pm\!20.07$     & $72.76\!\pm\!15.99$      & $74.35\!\pm\!17.43$      & $82.24\!\pm\!10.59$      & $82.93\!\pm\!19.29$ \\
    \midrule
    \multirow{2}{*}{\textbf{YOTO}}
     & clear    & $\mathbf{4.88\!\pm\!0.23}$   & $\mathbf{4.27\!\pm\!0.41}$   & $\mathbf{2.77\!\pm\!0.22}$   & $\mathbf{1.74\!\pm\!0.33}$   & $\mathbf{3.42\!\pm\!1.30}$ \\
     & occluded & $\mathbf{4.85\!\pm\!0.30}$   & $\mathbf{4.65\!\pm\!0.70}$   & $\mathbf{2.80\!\pm\!0.19}$   & $\mathbf{1.76\!\pm\!0.41}$   & $\mathbf{3.52\!\pm\!1.38}$ \\
    \bottomrule
    \end{tabular}}
    
    \vspace{4pt}
    \resizebox{\textwidth}{!}{
    \begin{tabular}{l l c c c c c}
    \toprule
    \multicolumn{7}{c}{\textbf{Rotation error (deg)}} \\
    \midrule
    Method & Condition & Drill & Squirrel & Monkey & Avocado & All \\
    \midrule
    \multirow{2}{*}{FoundationPose}
     & clear    & $5.84\!\pm\!0.57$        & $6.04\!\pm\!1.19$        & $4.32\!\pm\!0.97$        & $5.57\!\pm\!0.50$        & $5.44\!\pm\!1.04$ \\
     & occluded & $87.81\!\pm\!21.72$      & $88.28\!\pm\!13.48$      & $73.44\!\pm\!11.14$      & $92.73\!\pm\!8.70$       & $85.57\!\pm\!15.33$ \\
    \midrule
    \multirow{2}{*}{\textbf{YOTO}}
     & clear    & $\mathbf{4.78\!\pm\!0.39}$   & $\mathbf{4.85\!\pm\!0.62}$   & $\mathbf{3.11\!\pm\!0.46}$   & $\mathbf{1.59\!\pm\!0.27}$   & $\mathbf{3.58\!\pm\!1.44}$ \\
     & occluded & $\mathbf{4.72\!\pm\!0.39}$   & $\mathbf{4.88\!\pm\!0.63}$   & $\mathbf{3.12\!\pm\!0.46}$   & $\mathbf{1.51\!\pm\!0.25}$   & $\mathbf{3.56\!\pm\!1.46}$ \\
    \bottomrule
    \end{tabular}}
    \end{table}
    
    Under a clear view both methods recover sub-centimeter translation 
    and few-degree rotation error, with YOTO outperforming FoundationPose 
    on every object. Under occlusion the gap widens sharply: 
    FoundationPose's errors grow by $15.1\times$ in translation and 
    $15.7\times$ in rotation, whereas YOTO is essentially unaffected 
    (Tab.~\ref{tab:exp2}), as neither tactile input nor calibrated sensor 
    poses depend on the camera. Fig.~\ref{fig:teaser} (occluded column) 
    shows FoundationPose's bounding box latching onto the hand while 
    YOTO's axes remain anchored to the drill. Within YOTO, the per-object spread reveals a complementary trend to 
    Sec.~\ref{sec:exp_localization}: the drill, best for localization in 
    isolation, is now the most challenging for pose recovery. Its 
    rotationally symmetric long axis forces a near-coaxial dual grasp 
    that under-constrains the rotation about that axis in the SVD 
    recovery, an issue that resurfaces in dynamic tracking 
    (Sec.~\ref{sec:exp_dynamic}). The avocado, with broadly curved 
    geometry admitting non-coaxial contact, gives the smallest pose error.

    \vspace{-10pt}
    \subsection{Dynamic Pose Tracking from a Single Contact}
    \label{sec:exp_dynamic}
    
    \begin{figure*}[!t]
      \centering
      \includegraphics[width=\linewidth]{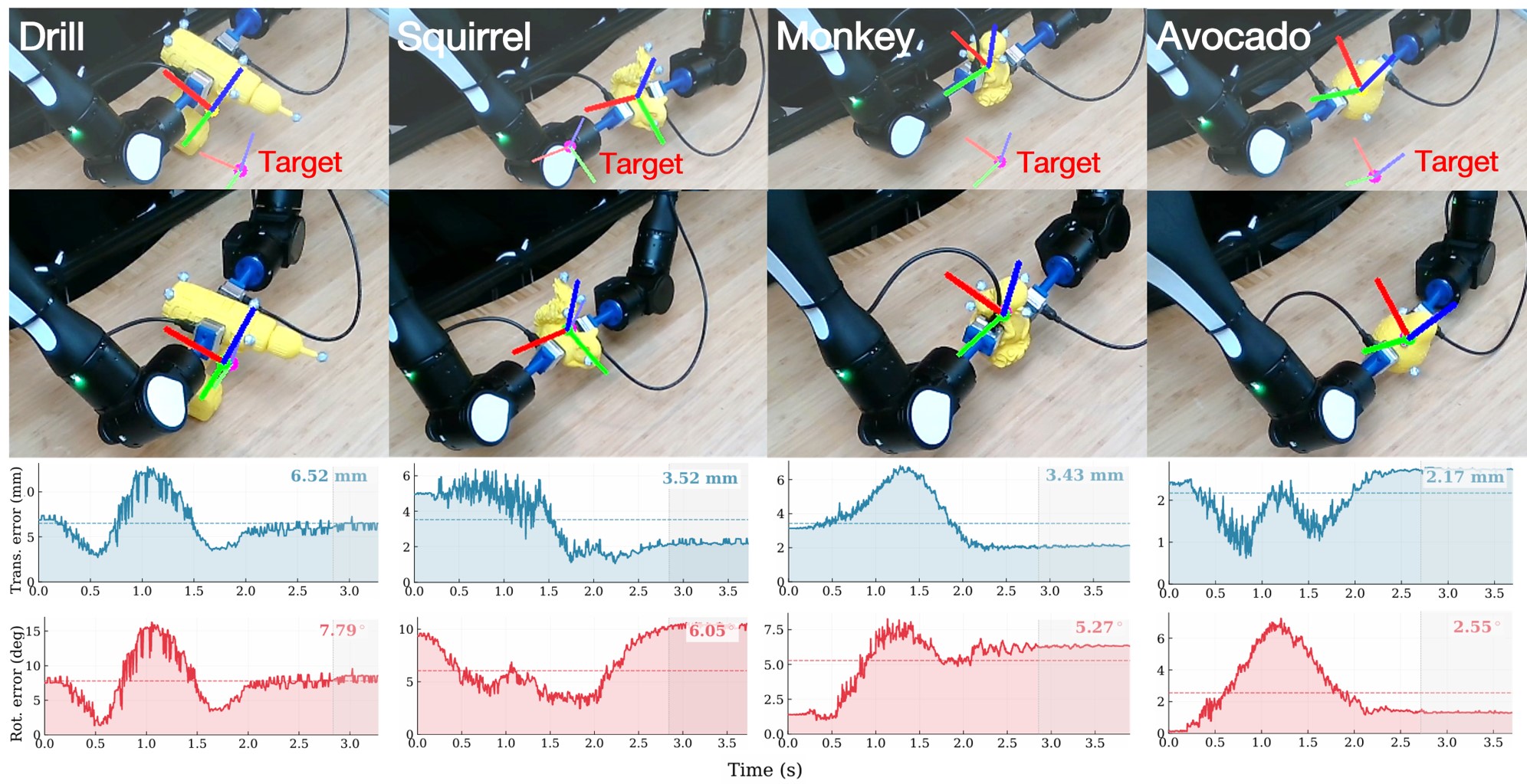}
      \captionsetup{font=footnotesize}
      \caption{Per-frame 6-DoF tracking error for a representative trajectory per object (top: translation; bottom: rotation). The dashed line in each panel marks the cross-trial mean during motion reported in Tab.~\ref{tab:exp3} (numeric value shown), provided as a reference overlay; the coloured curve illustrates the temporal shape of one representative trial. Shaded tail = post-motion settle.}
      \label{fig:exp3_curves}
      \vspace{-15pt}
    \end{figure*}
    
    Static placement-level accuracy (Sec.~\ref{sec:exp_pose}) does not 
    guarantee that the recovered pose remains valid during manipulation, 
    since soft-gel deformation, micro-slip, and inertial loading can 
    erode the rigid-grasp assumption. For each object, both gels 
    establish contact once, after which YOTO estimates the grasp 
    configuration and the system executes a randomly sampled dual-arm 
    SE(3) trajectory ($\sim$100\,mm translation, 
    $\sim$10--15$^\circ$ rotation) in open loop: no further sensing is 
    used and the pose is recovered analytically from live gel poses. 

    \begin{table}[!t]
    \vspace{-0pt}
    \centering
    \captionsetup{font=footnotesize}
    \caption{Dynamic tracking error (mean $\pm$ std over 5 trials per object, lower is better).}
    \label{tab:exp3}
    \small
    \setlength{\tabcolsep}{5pt}
    \begin{tabular}{l c c c c}
    \toprule
                        & Drill       & Squirrel    & Monkey      & Avocado \\
    \midrule
    Trans. (mm)         & $6.5\pm0.8$ & $3.5\pm0.4$ & $3.4\pm0.5$ & $2.2\pm0.3$ \\
    Rot.\ (\textdegree) & $7.8\pm1.1$ & $6.1\pm0.7$ & $5.3\pm0.6$ & $2.6\pm0.4$ \\
    \bottomrule
    \end{tabular}
    \vspace{-15pt}
    \end{table}
    
    Tab.~\ref{tab:exp3} reports mean tracking error across five trajectories per object. Mean translation error stays below 7\,mm and rotation error below 8$^\circ$ across all objects. Drill incurs the largest error, consistent with its elongated body and offset, unstable center of mass: rotational inertia and gravity-induced torque about the grasp axis amplify slip at the gel--object interface during motion. The other three objects, whose mass distributions better align with the grasp axis, retain near-static accuracy.
    
    Each curve in Fig.~\ref{fig:exp3_curves} shows a transient mid-motion 
    peak (max 7--16$^\circ$) followed by a settled tail. Peaks arise from 
    inertial loading and dual-arm asynchrony but remain bounded: at peak 
    excitation the tracker never diverges. The settled tail reflects 
    plastic grasp drift, which is small (1--3\,mm / 1--2$^\circ$) 
    and predictable from object geometry, distinguishing YOTO from 
    feedback-dependent trackers whose error typically diverges when 
    feedback is lost. Since YOTO is invoked only once per trajectory, the 
    bounded error serves as an end-to-end validation that a single touch 
    suffices for sustained 6-DoF tracking.


\section{Conclusion and Limitations}
\label{sec:conclusion}

    We presented YOTO, a tactile-only system that recovers absolute 6-DoF 
    object pose from a single simultaneous dual-GelSight contact. By 
    representing tactile patches as 3D point clouds and matching them to 
    a surface-block decomposition of the object, YOTO localizes each 
    contact and recovers the pose analytically through a normal-aware 
    SVD solver. A two-stage virtual-then-real training pipeline keeps 
    physical data requirements modest, and the system operates on 
    consumer-grade scanned models in addition to CAD meshes. Across four 
    geometrically diverse objects, YOTO attains a mean contact 
    localization error below $5\,$mm, sub-$2\,$mm / sub-$2^\circ$ pose 
    error on the avocado under heavy occlusion, and supports sustained dynamic tracking initialized from a single dual-contact touch with bounded error.

    These results suggest that tactile-only 6-DoF pose estimation is
    a viable alternative to vision-based pose in occlusion-heavy
    manipulation pipelines, rather than only a fallback when vision
    fails. The bottleneck for tactile perception is no longer the
    raw sensing resolution but the geometric representation that
    couples local contact patches to a known object surface. The surface-block decomposition introduced here is one such representation; extending it to handle larger object libraries with a shared, sensor-agnostic tactile encoder~\cite{zhao2024transferable,feng2025anytouch}, and to articulated or deformable objects whose state cannot be captured by a single rigid 6-DoF pose, are natural next steps.  
    
    \paragraph{Limitations.}
    YOTO still requires a per-object fine-tuning step with a small set of 
    real GelSight contacts (20 per object for fine-tuning, separate from 
    the 10 evaluation contacts in Sec.~\ref{sec:exp_localization}); fully 
    sample-free deployment remains future work. The SVD solver also degrades on near-coaxial grasps, which is unavoidable for rotationally symmetric objects like the drill. Finally, dynamic tracking rests on a rigid-grasp assumption that holds for centimeter-scale trajectories but accumulates plastic drift over longer manipulations. A closed-loop variant that re-runs the localisation network during motion would in principle mitigate this, since per-contact inference is fast enough (Tab.~\ref{tab:runtime}); the harder open problem is reconstructing a clean tactile patch from a gel that is continuously sliding and deforming under inertia, which we leave to future work.



\clearpage
\bibliography{example}  

\clearpage
\appendix

\section{Dataset and Object Models}
\label{app:dataset}

This section details how YOTO acquires per-object surface representations and
constructs the virtual tactile patch dataset used in the first stage of
training (Sec.~\ref{sec:training_objective}). The pipeline takes a
consumer-grade 3D scan as input and produces (i) a parent surface point cloud
with per-point normals, (ii) a partition of this cloud into surface blocks
aligned with the object's principal axes, and (iii) a large pool of virtual
tactile patches with ground-truth contact labels. All four evaluation objects (avocado, drill, squirrel, and monkey) are processed by the same pipeline; the default input throughout this section is the scanned mesh of Sec.~\ref{app:dataset_scan}, while for the YOTO (CAD mesh) ablation row of Tab.~\ref{tab:exp1} the input mesh is instead the original CAD source file from which the corresponding object was 3D-printed, with every subsequent step of this section (block decomposition, virtual patch sampling, labelling) remaining identical.

\subsection{Object Models and Scan Pipeline}
\label{app:dataset_scan}

Each object is scanned with the free \textbf{KIRI Engine} mobile application
running on a consumer tablet (iPad Pro; an iPhone with comparable rear-camera
hardware is also supported). The scan is performed handheld, sweeping the
device around the stationary object at a working distance of roughly
$20$--$30\,$cm. KIRI Engine's cloud reconstruction returns a watertight
triangulated mesh with per-vertex colour. We deliberately use a commodity mobile-photogrammetry pipeline, rather than a structured-light or laser scanner, so that the data-acquisition cost matches what an end user could
reproduce. The mesh is uniformly resampled into a parent point cloud
$\mathcal{P}=\{(\mathbf{x}_j,\mathbf{n}_j)\}_{j=1}^{N}$, with per-point
normals estimated from a 30-nearest-neighbour patch via Open3D's PCA
estimator, followed by a tangent-plane consistency pass to orient them
outward. Resulting parent-cloud sizes and bounding-box dimensions per object
are listed in Tab.~\ref{tab:per_obj}.

\begin{figure}[h]
    \centering
    \includegraphics[width=0.8\linewidth]{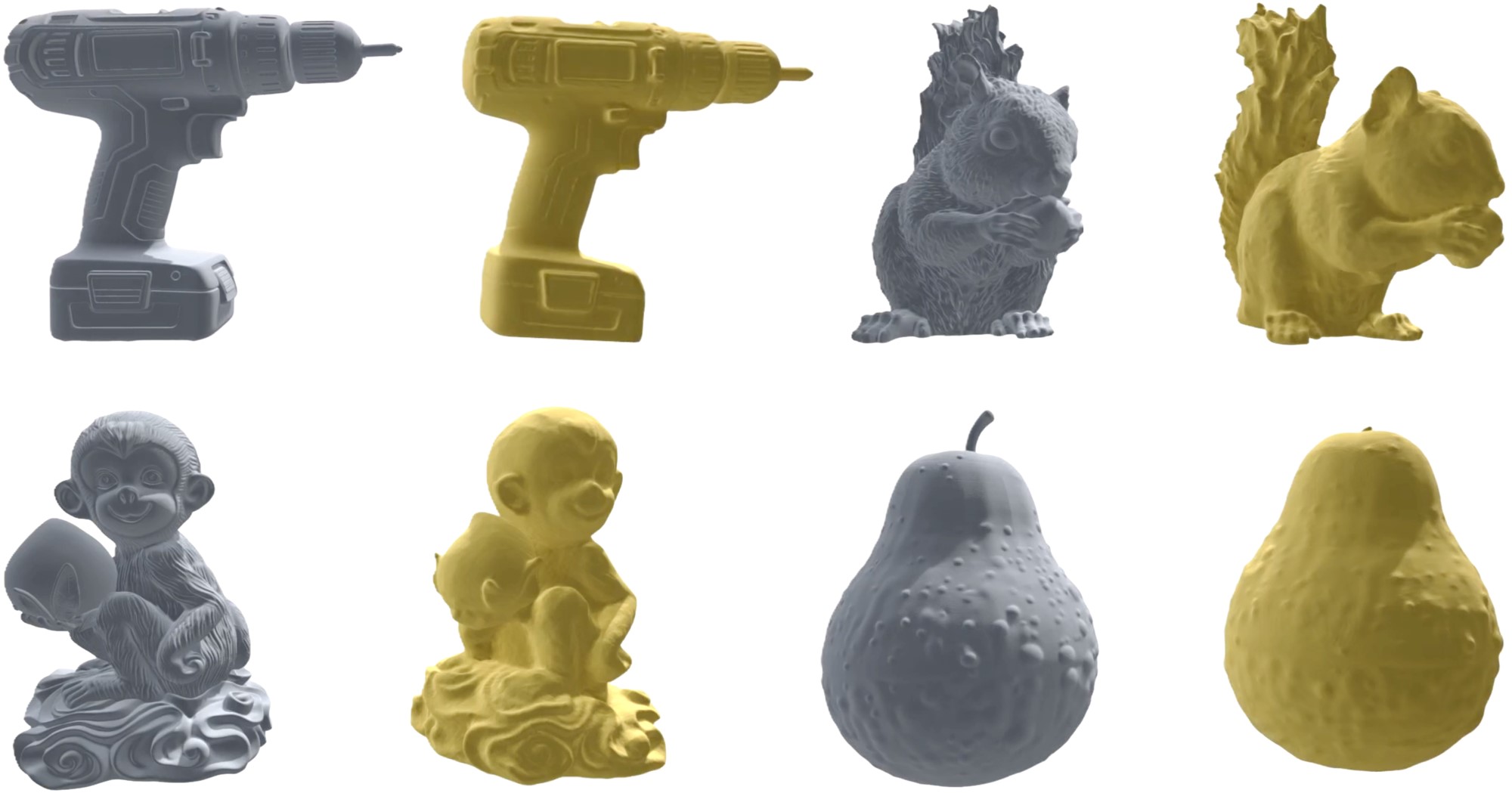}
    \captionsetup{font=footnotesize}
    \caption{CAD source meshes (grey) versus KIRI Engine mobile scans
    (yellow) for the four evaluation objects. Top row: drill and
    squirrel; bottom row: monkey and avocado. Within each pair, the
    CAD reference is on the left and the consumer-grade scanned
    reconstruction is on the right, rendered from the same viewpoint.}
    \label{fig:cad_vs_scan}
    \vspace{-5pt}
\end{figure}

\subsection{Surface Block Decomposition}
\label{app:dataset_blocks}

\paragraph{Principal-axis alignment.}
For each parent cloud we estimate a canonical object frame by PCA. The
eigenvector of the smallest eigenvalue of the centred covariance matrix is
taken as the up-direction $\mathbf{e}_z^O$, sign-aligned with world
positive $z$. The eigenvector of the largest eigenvalue is taken as
$\mathbf{e}_x^O$, and $\mathbf{e}_y^O = \mathbf{e}_z^O \times \mathbf{e}_x^O$
closes a right-handed frame. All downstream geometry (block partitioning, patch labels, and per-patch standardisation) is expressed in this object
frame.

\paragraph{Grid partition.}
Within the canonical frame we partition $\mathcal{P}$ on a regular
$6\times3\times3$ grid (54 nominal blocks per object), giving each block
$\mathcal{B}_i$ an object-frame centre $\mathbf{c}_i^O$ and a fixed point
membership. Blocks holding fewer than 10 points are discarded as numerically
unstable. As Tab.~\ref{tab:per_obj} shows, the four objects retain between
43 and 51 valid blocks; the drill loses the most cells because its elongated,
narrow profile leaves several grid cells empty.

\paragraph{Per-block standardisation.}
For each valid block we estimate a local outward-normal direction
$\bar{\mathbf{n}}_i$ by averaging the block's point normals and verifying
that $\bar{\mathbf{n}}_i$ points away from the global object centre
(sign-flipped otherwise). The block's \emph{concave} direction is taken as
$-\bar{\mathbf{n}}_i$, and a rotation $\mathbf{R}_i\in SO(3)$ is computed via
Rodrigues' formula that maps this concave direction onto positive $z$. The
pre-rotated block point cloud, together with $\mathbf{R}_i$ and
$\mathbf{c}_i^O$, is cached and reused as the retrieval key against which
incoming tactile patches are matched.

\subsection{Virtual Tactile Patch Generation}
\label{app:dataset_virtual}

\paragraph{Parent rotation augmentation.}
To prevent the localisation network from memorising a single canonical
alignment of the parent cloud, we generate \textbf{29 parent variants per
object}: the unrotated original, 14 evenly-spaced fixed rotations at
$12^{\circ}$ increments spanning $[12^{\circ},168^{\circ}]$, and 14
randomly-sampled rotations drawn uniformly from $[0^{\circ},360^{\circ})$
and filtered to be at least $0.5^{\circ}$ from any fixed angle. All
rotations are applied about the canonical $\mathbf{e}_z^O$ axis. The block
decomposition is rotated rigidly with each variant; block indices and
canonical centres remain consistent across variants so that ground-truth
labels transfer without re-computation.

\paragraph{Patch sampling and acceptance criteria.}
For each parent variant we attempt to draw 300 virtual tactile patches.
Each attempt samples a candidate centre uniformly inside the safe interior
of the parent cloud (no closer than $1.2\,r$ to any bounding-box face, where
$r$ is the current trial radius). Around each centre we extract all parent
points within a Euclidean ball, with the ball radius binary-searched over
$[0,\,r]$ ($r=20\,$mm, comparable to the GelSight Mini contact footprint)
until the enclosed point count lands in the target range $[700,1200]$. If
the binary search fails to converge inside this band, or if the resulting
patch yields a degenerate concave direction (i.e.\ rotated-frame
$z$-component below $0.99$), the candidate is rejected and the attempt is
retried.

\paragraph{Ground-truth labelling.}
Each accepted patch is converted to the canonical form expected by the
localisation network at inference time:
\begin{enumerate}
    \item the patch centroid is offset to the in-cloud point closest to
          it, giving the object-frame contact location $\mathbf{p}^O$;
    \item the \emph{dominant} block $i^\star$, defined as the valid block           holding the largest fraction of the patch's points, is recorded as        the coarse-stage retrieval target;
    \item the residual $\Delta\mathbf{p}^O = \mathbf{p}^O -
          \mathbf{c}_{i^\star}^O$ is recorded as the fine-stage regression
          target;
    \item the patch's concave direction is estimated from PCA
          (smallest-eigenvalue eigenvector, sign-disambiguated by comparing
          centre-vs-edge height along that direction), and the patch
          coordinates and normals are rotated so this direction aligns
          with positive $z$ axis.
\end{enumerate}

\paragraph{Why PCA for patches but normals for blocks?}
Block standardisation uses averaged surface normals because the parent
cloud's normals are reliable (mesh-derived). Patches, on the other hand,
must use the same standardisation procedure at inference time, when the
input is a real GelSight reconstruction whose absolute normals are noisier.
The PCA-based concave direction is more robust to this noise and gives a
consistent standardisation between virtual training patches and real-tactile
inputs.

\subsection{Per-Object Statistics}
\label{app:dataset_stats}

    \begin{table}[h]
    \centering
    \captionsetup{font=footnotesize}
    \caption{Per-object dataset statistics. ``Parent cloud $N$'' is the number of
    points sampled from the scanned mesh. ``Bbox'' reports the axis-aligned
    bounding box in the canonical object frame, in millimetres. ``Valid blocks''
    is the number of non-empty cells of the $6{\times}3{\times}3$ grid after the
    $\geq 10$-point threshold. ``Virtual patches'' is the number of accepted
    patches after point-count and concave-direction filtering, out of a nominal
    budget of $29\times300=8{,}700$ per object.}
    \label{tab:per_obj}
    \small
    \setlength{\tabcolsep}{8pt}
    \begin{tabular}{l r c c r}
    \toprule
    Object   & Parent cloud $N$ & Bbox (mm$^3$)               & Valid blocks & Virtual patches \\
    \midrule
    Avocado  & $72{,}138$ & $73.6 \times 71.7 \times 91.0$  & $50\,/\,54$ & $8{,}700$ \\
    Drill    & $74{,}631$ & $175.6 \times 24.1 \times 153.3$& $43\,/\,54$ & $8{,}686$ \\
    Squirrel & $43{,}887$ & $58.1 \times 102.2 \times 115.0$& $49\,/\,54$ & $8{,}700$ \\
    Monkey   & $93{,}041$ & $73.4 \times 62.5 \times 110.0$ & $51\,/\,54$ & $8{,}700$ \\
    \midrule
    Total    & ---        & ---                              & ---         & $34{,}786$ \\
    \bottomrule
    \end{tabular}
    \end{table}

The drill's elongated, narrow profile is reflected in both its lower
valid-block count ($43$, vs.\ $49$--$51$ for the other three objects) and in
the small number of patches ($14$) rejected at sampling time, where the
radius binary-search cannot reconcile the $[700,1200]$ point-count band with
the object's $24\,$mm narrow axis. Across all four objects the pipeline
yields $34{,}786$ virtual tactile patches, which constitute the entirety of
the Stage-1 (virtual-pretraining) supervision in
Sec.~\ref{sec:training_objective}. Real GelSight data used in Stage 2 is
described in Sec.~\ref{app:data_collection}.

\section{Real GelSight Data Collection Protocol}
\label{app:data_collection}

This section describes how the real-world GelSight contacts used for
Stage-2 fine-tuning and for the quantitative localisation evaluation
of Sec.~\ref{sec:exp_localization} are acquired and converted into
ground-truth-labelled training samples. The downstream pose-estimation
and dynamic-tracking experiments of Sec.~\ref{sec:exp_pose} and
Sec.~\ref{sec:exp_dynamic} use the AIRBOT Play dual-arm setup
described there. The present section covers only the data-collection
step that produces real contacts for fine-tuning and for the
Sec.~\ref{sec:exp_localization} evaluation. That step requires no
robotic manipulator. Contacts are produced by hand, and ground-truth
contact locations come entirely from motion-capture geometry rather
than from arm forward kinematics.

\subsection{Sensor Rig and Mocap Setup}
\label{app:dc_hardware}

\begin{figure}[h]
    \centering
    \includegraphics[width=0.55\linewidth]{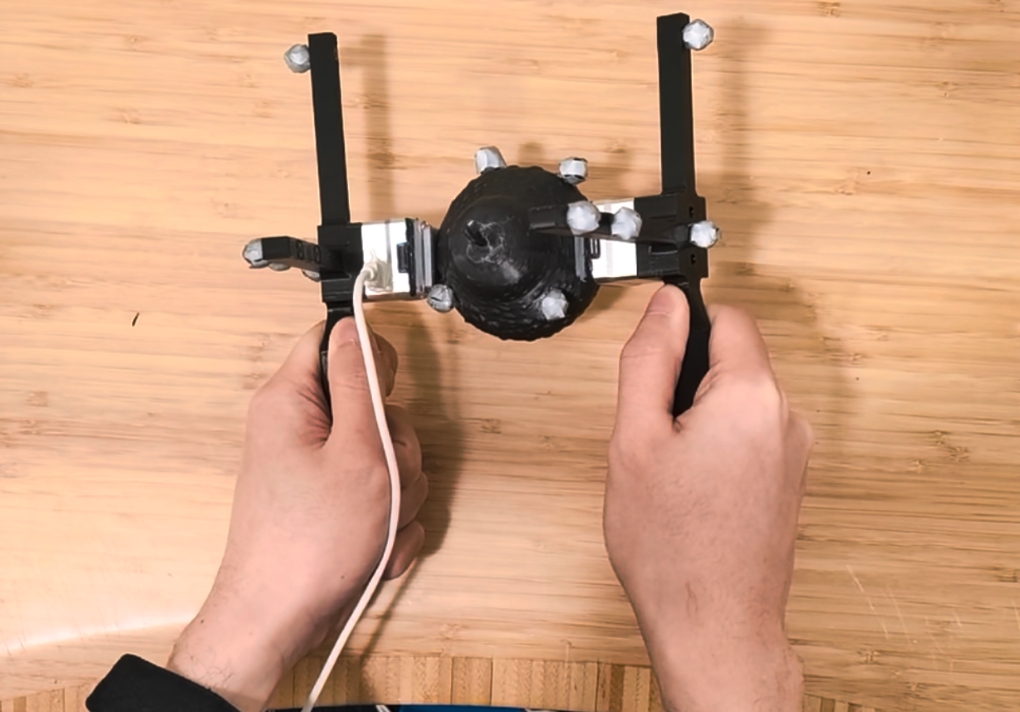}
    \captionsetup{font=footnotesize}
    \caption{The pair of 3D-printed L-shaped handheld rigs used for
    real-world data collection (Sec.~\ref{app:dc_hardware}). Each rig
    holds one GelSight Mini and carries four passive retro-reflective
    markers forming its OptiTrack rigid body. The photograph shows the
    two rigs during the initial setup; data collection itself uses one
    rig at a time and is performed single-handed.}
    \label{fig:rig}
    \vspace{-5pt}
\end{figure}

Data is collected with a pair of identical L-shaped 3D-printed
handheld rigs, one per GelSight Mini sensor, matching the left/right
dual-sensor configuration used throughout the main paper. Each rig
holds a single GelSight Mini at one end and carries four passive
retro-reflective markers distributed across its arms; together these
four markers define an OptiTrack rigid body whose pose is streamed
live. The two rigs are calibrated independently (see
Sec.~\ref{app:dc_calibration}) and the resulting left/right
calibrations are stored as separate files. The object under test is
itself instrumented with four to five small retro-reflective beads
($5\,$mm radius) adhered directly to its surface at non-coplanar
positions, defining a third rigid body. All rigid bodies are tracked by an
OptiTrack motion-capture system. The whole setup fits on a desktop
and is operated by one experimenter, who brings a single rig into
contact with the stationary object by hand for each real-data sample
while the GelSight RGB stream and the relevant rigid-body poses are
recorded into a synchronised ROS bag. Fig.~\ref{fig:rig} shows the
pair of rigs side-by-side.

\vspace{+10pt}

\subsection{Calibration Chain}
\label{app:dc_calibration}

The pipeline requires two independent calibrations, both performed
once per physical setup, followed by a per-trial composition that
yields the GelSight surface centre in the object frame.

\paragraph{(i) GelSight-surface centre w.r.t.\ the rig rigid body.}
The GelSight gel face is a soft hemisphere whose centre cannot be
directly observed by mocap. We recover its location
$\mathbf{p}^R_{gel}$ in the rig's rigid-body frame via a multi-pose
dual-ball procedure. A small reflective ball ($5\,$mm) is temporarily
fixed at the geometric centre of the gel face, and a second reference
ball is mounted on the rig at a non-collinear offset. With both balls
visible to the mocap system, the experimenter moves the rig through
$5$ distinct poses; in each pose the two ball centres are recorded in
the mocap frame, mapped back into the rig's rigid-body frame via the
streamed rigid-body pose, and averaged across poses. The gel-ball
position in the rig frame stays within sub-millimetre dispersion
across the five poses, confirming both the mocap accuracy and the
rigidity of the rig assembly.

\paragraph{(ii) STL model frame w.r.t.\ the object rigid body.}
For each object we pick the four to five non-coplanar surface points
on the scanned mesh (Sec.~\ref{app:dataset_scan}) at the locations
of the adhered reflective beads, using an interactive 3D viewer that
snaps clicks to the nearest surface vertex and offsets along the
local outward normal by the bead radius. This gives a set of ball-centre positions $\{\mathbf{P}^M_i\}$ in the
object (STL) frame. We then record the same ball centres
$\{\mathbf{P}^W_i\}$ in the mocap frame and
solve the rigid alignment $\mathbf{T}^W_M$ in closed form via SVD
(orthogonal Procrustes). On a representative object (avocado), the
per-marker SVD residuals are $\{1.52,\,1.67,\,1.82,\,2.23\}\,$mm,
with a mean of $1.81\,$mm and a maximum of $2.23\,$mm. We
additionally record the object's rigid-body pose
$\mathbf{T}^W_{R_{obj}}$ at this instant, and store the fixed offset
$\mathbf{T}^{R_{obj}}_M = (\mathbf{T}^W_{R_{obj}})^{-1}\mathbf{T}^W_M$,
which allows the object to be freely repositioned between contacts
without re-calibration.

\paragraph{(iii) Per-frame composition.}
At runtime, the live rig pose $\mathbf{T}^W_{R_{gel}}(t)$ and live
object rigid-body pose $\mathbf{T}^W_{R_{obj}}(t)$ are combined with
the two static offsets above to map the gel-surface centre into the
object frame,
\begin{equation}
\mathbf{p}^M_{gel}(t)
\;=\;
\bigl(\mathbf{T}^W_{R_{obj}}(t)\,\mathbf{T}^{R_{obj}}_M\bigr)^{-1}
\;\mathbf{T}^W_{R_{gel}}(t)\;
\mathbf{p}^R_{gel}.
\label{eq:gel_in_model}
\end{equation}
Every quantity on the right-hand side is either fixed by calibration
or read from mocap; no kinematic chain enters the formula.

\subsection{Tactile Contact Trial Acquisition}
\label{app:dc_trial}

For each contact the experimenter brings the rig into static contact
with the object for a short hold ($\approx 1\,$s), then withdraws.
During the trial, the ROS bag records (i) the rig rigid-body pose,
(ii) the object rigid-body pose, and (iii) the GelSight RGB image
stream, all time-stamped to the same clock. The GelSight image
stream is converted to a local 3D point cloud using the
InvariantCloud reconstruction pipeline~\cite{ye2026invariantcloud},
yielding one tactile point cloud per contact in the sensor frame.
Only the bag-level synchronisation between mocap and image streams is
needed; image-to-mocap intrinsic calibration is not required because
the tactile reconstruction is expressed in the sensor frame, not in
the camera frame.

\subsection{Ground-Truth Contact Point Extraction}
\label{app:dc_gt}

Eq.~\ref{eq:gel_in_model} is applied to every frame in the bag to
obtain the gel-surface centre's trajectory in the object frame,
$\{\mathbf{p}^M_{gel}(t)\}$. The scanned mesh is loaded into a
ray-casting scene and the signed distance from each
$\mathbf{p}^M_{gel}(t)$ to the mesh surface is computed; frames with
a non-positive signed distance are classified as \emph{contact
frames}, since the soft gel slightly penetrates the rigid mesh model
on real contact. The closest mesh point of each contact frame is
taken as a per-frame estimate of the true contact location; these
estimates are aggregated by averaging, with a $3\sigma$ outlier
rejection on Euclidean deviation from the mean, to produce the final
ground-truth contact location $\mathbf{p}^O$ in the object frame.

The tactile point cloud (Sec.~\ref{app:dc_trial}) is then
standardised to the same canonical form as the virtual patches
(Sec.~\ref{app:dataset_virtual}). PCA-based concave-direction
estimation rotates the patch so the surface concavity points along
the positive $z$ axis, and the patch is centred at its centroid. The
$\mathbf{p}^O$ recovered above is paired with this standardised patch
to compute the ground-truth dominant block $i^\star$ and offset
$\Delta\mathbf{p}^O$ exactly as in the virtual pipeline. The
resulting record, comprising patch coordinates, normals, contact
location, dominant block, and offset, is saved as a single training
sample.

\subsection{Per-Object Real Data Yield}
\label{app:dc_yield}

Per object, 30 real GelSight contacts are collected at locations
distributed across the surface. Of these, 20 are used for
Stage-2 fine-tuning (Sec.~\ref{sec:training_objective}) and the
remaining 10 are held out as the evaluation set reported in
Sec.~\ref{sec:exp_localization}. Across the four objects this yields
$120$ labelled real tactile samples in total. The fine-tune split is
seen jointly with the virtual patches of
Sec.~\ref{app:dataset_virtual} during Stage-2 training, with a
larger per-sample loss weight to counter the virtual-versus-real
imbalance.

\section{Implementation Details}
\label{app:implementation}

\subsection{Network Architecture}
\label{app:impl_arch}

\paragraph{Shared point-cloud encoder.}
The tactile patch and the surface blocks are encoded by the same
lightweight three-stage point-cloud module
(Sec.~\ref{sec:localization_network}). A point-wise MLP first lifts
the $6$-D input (coordinates + normals) to a $256$-D hidden feature.
Multi-scale local aggregation with $k\in\{8,16,32,64\}$ nearest
neighbours then runs, per scale, an MLP over the concatenation of
neighbour features and relative coordinates followed by max-pool
over neighbours. A fusion MLP concatenates the point-wise feature
with the four scale features and projects to a $256$-D hidden,
followed by a $256{\to}512$ pre-pool projection and max+mean
pooling to a single $512$-D vector. At training time $512$ points
are sub-sampled per patch or block, with repeat-padding when fewer
points are available. Encoder weights are shared between the
tactile and block branches.

\paragraph{Block matching head.}
The tactile feature $\mathbf{q}_s$ is matched against all valid
blocks of the queried object by cosine similarity scaled by a
learnable temperature $\tau$ (initialised at $10$). The
highest-scoring block defines the coarse hypothesis; its
$K_{\text{nbr}}{=}9$ spatial neighbours, by Euclidean distance over
block centres, form the candidate pool, which the fine head scores
and ranks by the combined coarse-plus-fine score
(Sec.~\ref{sec:localization_network}).

\paragraph{Fine prediction head.}
For each candidate block, the corresponding block feature is
concatenated with the tactile feature $\mathbf{q}_s$ and passed
through a two-layer MLP ($\cdot\!\to\!512\!\to\!256$, LayerNorm,
GELU, dropout $0.2$). Two linear heads emit the $3$-D offset
$\Delta\hat{\mathbf{p}}_s^O$ and a scalar confidence; the final
candidate selection takes the argmax of the coarse cosine score
plus this confidence.

\subsection{Training Schedule}
\label{app:impl_training}

Both stages use AdamW with weight decay $0.01$ and a OneCycleLR
schedule (cosine annealing, $10\%$ warm-up, div-factor $10$,
final-div-factor $1000$). All training is mixed-precision
(\texttt{bfloat16}) on a single GPU with batch size $32$.

\paragraph{Stage 1 (virtual pretraining).}
$500$ epochs over the virtual patch pool of
Sec.~\ref{app:dataset_virtual}. Base learning rate
$2\!\times\!10^{-3}$, with a $20\times$ lower rate
($1\!\times\!10^{-4}$) on the shared encoder parameters to
stabilise early training. Patch augmentations include random
in-plane rotation around the canonicalised patch's $z$ axis
($\pm 1\,$rad, applied with probability $0.75$), Gaussian jitter
($\sigma{=}0.01$, clipped at $0.05$), random point dropout (ratio
$0.1$, applied to $50\%$ of patches), and voxel down-sampling on a
$0.2\,$mm grid. These per-step patch augmentations are layered on
top of the static $29$-variant parent-rotation augmentation of
Sec.~\ref{app:dataset_virtual}.

\paragraph{Stage 2 (real fine-tuning).}
$400$ epochs over a mixed pool of virtual and real samples. A
custom batch sampler over-samples real contacts by a factor of
$10$ so that every mini-batch contains both modalities despite the
small real-data pool; in addition, the per-sample loss carries a
weight of $2.0$ on real samples relative to virtual samples, so
that each real contact contributes twice as much gradient as a
virtual one of equivalent loss. The base learning rate drops to
$4\!\times\!10^{-4}$ (shared encoder $1\!\times\!10^{-5}$).
Augmentations on the real side match Stage~1.

\paragraph{Loss weights.}
The coarse-stage classification loss
($\mathcal{L}_{\mathrm{cls}}$) and the two top-$K$ margin terms
($\mathcal{L}_{\mathrm{topK}}$, Sec.~\ref{sec:training_objective})
are combined with weight $\lambda_c{=}1.0$, and the fine-stage
smooth-$\ell_1$ position loss ($\mathcal{L}_{\mathrm{pos}}$) is
weighted at $\lambda_p{=}2.0$. The classification weight rises
adaptively (up to $2.5\times$) when top-$K$ recall is low, keeping
the coarse stage competitive on hard batches.

\subsection{Inference}
\label{app:impl_inference}

At test time the encoder features of all valid blocks for the four
object models are pre-computed once and cached; only the tactile
patch needs to be processed live. A single forward pass is
dominated by the tactile encoder over the $700$--$1200$-point patch
plus a cosine-similarity scan against the cached blocks (one
matrix-vector product per object). The recovered object-frame
contact $\hat{\mathbf{p}}_s^O$ is passed unchanged to the
normal-aware SVD solver of Sec.~\ref{sec:pose_recovery}.

\paragraph{Wall-clock runtime.}
Tab.~\ref{tab:runtime} compares per-contact inference latency
for YOTO and for the multi-start ICP baseline of
Sec.~\ref{sec:experiments}, measured on the same workstation
(NVIDIA RTX~4080 Laptop GPU with $12\,$GB VRAM for YOTO;
Intel Core i9 14th-generation CPU for the multi-threaded Open3D
ICP), on the same $24$ real evaluation contacts per object as
Sec.~\ref{sec:exp_localization}. YOTO timings are end-to-end
wall-clock measurements taken with GPU synchronisation
before and after each inference call, with the first call
excluded as warm-up.

\begin{table}[h]
\centering
\captionsetup{font=footnotesize}
\caption{Per-contact inference latency on the four test objects.
YOTO (this work) versus multi-start ICP. Each value is the mean
over $24$ real GelSight contacts per object; \emph{Speedup} is
the per-object ratio of ICP time over YOTO time.}
\label{tab:runtime}
\small
\setlength{\tabcolsep}{8pt}
\begin{tabular}{l r r r}
\toprule
Object   & YOTO (ms) & ICP (ms) & Speedup \\
\midrule
Drill    & $25.74$ & $580$ & $22.5\times$ \\
Squirrel & $18.30$ & $830$ & $45.4\times$ \\
Monkey   & $18.45$ & $770$ & $41.7\times$ \\
Avocado  & $26.50$ & $580$ & $21.9\times$ \\
\midrule
Mean     & $\mathbf{22.24}$ & $\mathbf{690}$ & $\mathbf{31.0\times}$ \\
\bottomrule
\end{tabular}
\end{table}

The mean $31\times$ speedup comes from two structural choices:
block features are pre-computed and cached, so per-contact
inference reduces to a single tactile-encoder forward pass plus
a cosine-similarity lookup against cached blocks, whereas ICP
must run a 30-start point-to-plane optimisation from scratch
per contact. A complementary property is the per-object spread:
YOTO's per-object mean latency varies by at most $8\,$ms across
the four objects while ICP varies by $250\,$ms, since ICP's
runtime grows with geometric complexity (more local minima per
parent surface) while YOTO's is bounded by the fixed-size
encoder forward pass.

\section{Failure Cases and Limitations}
\label{app:limitations}

This appendix expands on the per-object failure modes briefly
noted in Sec.~\ref{sec:experiments} and records two further
pipeline-level limitations beyond those listed in
Sec.~\ref{sec:conclusion}.

\subsection{Failure Modes by Object Geometry}
\label{app:fail_geom}

\paragraph{Wrong-block coarse retrieval (avocado).}
The avocado's surface bumps are mutually similar over
patch-sized neighbourhoods, so distinct $20\,$mm patches from
different bumps produce encoder features close to one another in
the matching space. On a minority of evaluation contacts the
coarse-stage cosine matcher latches onto a wrong block, and
because the fine head regresses an offset relative to that
block's centre rather than recovering global structure, the
resulting prediction is biased by the inter-bump separation
rather than by small local error. This is the dominant
contributor to the avocado's higher localisation mean in
Tab.~\ref{tab:exp1} and to its $5$--$10\,$mm tail in the
per-contact error distribution.

\paragraph{Coaxial-grasp rotation ambiguity (drill).}
For elongated, rotationally near-symmetric objects, the two
sensors' chosen contact points lie close to the object's long
axis on opposite sides of the body. The contact-to-contact
direction $\mathbf{d}^O$ then aligns with the axis of symmetry,
and the two estimated surface normals are anti-parallel along
that same axis. The three columns of $\mathbf{Q}^O$ in
Sec.~\ref{sec:pose_recovery} consequently span only a
two-dimensional subspace, leaving rotation about the long axis
under-determined. The SVD solver still returns a single answer,
but its component about the symmetry axis is dominated by
feature noise rather than geometry, which explains why the drill
incurs the largest rotation errors in Tab.~\ref{tab:exp2} and
Tab.~\ref{tab:exp3} despite its lowest localisation error in
Tab.~\ref{tab:exp1}.

\paragraph{Plastic drift over long open-loop motions.}
The dynamic-tracking results of Sec.~\ref{sec:exp_dynamic}
assume rigid gel--object contact. In practice the gel exhibits
small plastic deformation under inertial loading, and the
contact slowly shifts on the object surface over the course of
a trajectory. The drift magnitude reported in the main paper
($1$--$3\,$mm / $1$--$2^{\circ}$ over $\sim\!100\,$mm
trajectories) is bounded enough that a single-touch
initialisation remains useful, but the drift grows roughly
monotonically with motion length and would eventually exceed
task tolerances on longer manipulations.

\subsection{Additional Pipeline-Level Limitations}
\label{app:fail_pipeline}

\paragraph{Scan-quality dependency.}
The surface-block representation is built directly from the
KIRI Engine scan (Sec.~\ref{app:dataset_scan}), so localisation
inherits the scan's geometric fidelity. Objects with regions
poorly recovered by mobile photogrammetry, such as thin
protrusions or highly specular and transparent surfaces, would
introduce a systematic bias that the network has no signal to
correct, and a quantitative drop relative to the four reported
objects should be expected on such cases.

\paragraph{Single-shot scope.}
By construction YOTO consumes exactly one simultaneous
dual-contact observation. A grasp that is mid-slip or only
barely engaged at the moment of acquisition produces a noisy
local point cloud and therefore a noisy contact estimate. The
current pipeline has no mechanism to detect and reject such
low-confidence inputs; a practical deployment would add a
re-touch policy that triggers a new acquisition when the
matching confidence falls below a threshold.

\end{document}